\documentclass[conference]{IEEEtran}
\IEEEoverridecommandlockouts

\usepackage{cite}
\usepackage{amsmath,amssymb,amsfonts}
\usepackage{algorithmic}
\usepackage{graphicx}
\usepackage{textcomp}
\usepackage{xcolor}
\usepackage{booktabs}
\usepackage{multirow}
\usepackage{url}
\usepackage{tikz}
\usepackage{pgfplots}
\pgfplotsset{compat=1.17}
\usetikzlibrary{arrows.meta, positioning, shapes.geometric}

\def\BibTeX{{\rm B\kern-.05em{\sc i\kern-.025em b}\kern-.08em
    T\kern-.1667em\lower.7ex\hbox{E}\kern-.125emX}}

\begin{document}

\title{Domain-Adapted Small Language Models with Hybrid Post-Processing: Achieving Cost-Efficient, Low-Latency Multi-Label Structured Prediction via LoRA Fine-Tuning on Scarce Data}

\author{
\IEEEauthorblockN{Srinivasan Manoharan}
\IEEEauthorblockA{PayPal Inc\\
San Jose, USA\\
srinivmanoharan@paypal.com}
\and
\IEEEauthorblockN{Dilipkumar Nallusamy}
\IEEEauthorblockA{PayPal Inc\\
Chennai, India\\
dnallusamy@paypal.com}
\and
\IEEEauthorblockN{Sachin Kumar}
\IEEEauthorblockA{PayPal Inc\\
Bangalore, India\\
skumar82@paypal.com}
\and
\IEEEauthorblockN{Haifeng Wu}
\IEEEauthorblockA{PayPal Inc\\
San Jose, USA\\
haifwu@paypal.com}
}

\maketitle

\begin{abstract}
Deploying frontier large language models (LLMs) for domain-specific structured evaluation tasks incurs prohibitive latency, cost, and data-privacy overhead. We present a hybrid framework that fine-tunes a small language model (LLaMA 3.1 8B, 2.05\% trainable parameters via LoRA) on only 219 curated examples and couples it with a deterministic rule-based post-processing layer. Applied to multi-label compliance evaluation of conversational transcripts (18 heterogeneous output fields), our system achieves 100\% JSON structural validity, 83.0\% human-validated overall accuracy, and 100\% accuracy on the most critical classification field in blind evaluation on 53 unseen production transcripts. On a single NVIDIA A100 GPU, inference completes in $\sim$2 seconds---2--5$\times$ faster than frontier APIs---at \$0.013 per evaluation versus \$0.025--\$0.055 for proprietary alternatives, yielding 46--76\% cost savings. We introduce targeted hard-negative augmentation for critical decision boundaries and formalize the hybrid neural-symbolic decomposition, demonstrating that domain-adapted small language models with post-processing can match frontier model accuracy while dramatically reducing operational cost, latency, and privacy risk.
\end{abstract}


\section{Introduction}
Organizations increasingly deploy AI-powered conversational agents in regulated domains such as financial services, healthcare, and telecommunications~\cite{brown2020}. These agents must comply with strict operational standards---identity disclosures, authorization protocols, and outcome classification---defined by Standard Operating Procedures (SOPs). Manual quality assurance is labor-intensive, subjective, and does not scale.

Frontier LLMs (GPT-4o, Claude Opus) offer strong general-purpose capabilities but present operational challenges for domain-specific structured evaluation: (1) high per-query costs at scale, (2) 4--10$\times$ higher latency than self-hosted models, (3) accuracy degradation when domain rules are embedded in large context windows~\cite{liu2024lost}, and (4) data privacy concerns when sensitive transcripts leave organizational boundaries.

We demonstrate that a \textbf{domain-adapted small language model} (8B parameters) with \textbf{hybrid post-processing} can match or exceed frontier model accuracy on structured compliance evaluation while providing order-of-magnitude improvements in cost and latency. Our contributions are:
\begin{enumerate}
    \item \textbf{Data-efficient domain adaptation}: Fine-tuning LLaMA 3.1 8B via LoRA~\cite{hu2022lora} with only 219 examples, including targeted hard-negative augmentation for critical decision boundaries.
    \item \textbf{Hybrid neural-deterministic inference}: A two-stage pipeline decomposing evaluation into contextual understanding (neural) and invariant rule enforcement (symbolic).
    \item \textbf{Production evaluation}: Blind assessment on 53 production transcripts validated by human experts, with detailed cost and latency analysis against frontier API alternatives.
\end{enumerate}

\section{Related Work}
\textbf{Parameter-efficient fine-tuning.} LoRA~\cite{hu2022lora} introduces low-rank weight decompositions, reducing trainable parameters by orders of magnitude. QLoRA~\cite{dettmers2023qlora} adds 4-bit quantization. These have been applied to legal~\cite{cui2023chatlaw}, medical~\cite{singhal2023clinical}, and financial~\cite{wu2023bloomberggpt} domains. We extend PEFT to multi-label compliance evaluation with extreme data scarcity.

\textbf{Long-context limitations.} Liu et al.~\cite{liu2024lost} show 20--40 point accuracy drops when relevant information is in the middle of long contexts, motivating weight-encoded domain knowledge over context-window approaches.

\textbf{Structured output generation.} Constrained decoding~\cite{willard2023guided} enforces syntactic validity but not semantic correctness. Our hybrid approach addresses both via fine-tuning and rule-based post-processing.

\textbf{Data augmentation.} We draw on contrastive learning principles~\cite{gao2021simcse} but operate at the input-output pair level, creating near-miss variants for critical compliance patterns.

\section{Problem Formulation}
\subsection{Task Definition}
Given a transcript $T = \{(s_i, u_i)\}_{i=1}^{n}$ and disposition $d$, produce a structured vector $\mathbf{y} = (y_1, \ldots, y_{18})$ where each $y_k \in \mathcal{Y}_k$. The 18 fields span four types:

\noindent\textbf{Type I (Critical):} Agent Disclosure $y_{\text{AD}} \in \{P, F\}$, Payment Disclosure $y_{\text{PD}} \in \{P, F, N/A\}$, Disposition $y_{\text{D}} \in \{P, F\}$, Compliance Result $y_{\text{CR}} \in \{P, F\}$.

\noindent\textbf{Type II (Behavioral):} Threats, Unprofessionalism, Transfer Accuracy, etc.

\noindent\textbf{Type III (Informational):} Product Name, Balance Due, Amount Collected.

\noindent\textbf{Type IV (Free-text):} Overall Feedback, Expected Responses.

\subsection{Inter-Field Constraints}
The Compliance Result is deterministically derived:
\begin{equation}
y_{\text{CR}} =
\begin{cases}
\text{Fail} & \text{if } y_{\text{AD}}{=}F \vee y_{\text{PD}}{=}F \vee y_{\text{D}}{=}F \\
\text{Pass} & \text{otherwise}
\end{cases}
\end{equation}

Agent Disclosure requires exact substring matching against a canonical phrase $\phi^{*}$:
\begin{equation}
y_{\text{AD}} =
\begin{cases}
\text{Pass} & \text{if } \phi^{*} \sqsubseteq T \\
\text{Fail} & \text{otherwise}
\end{cases}
\end{equation}

\section{Methodology}
\subsection{System Architecture}
\begin{figure}[t]
\centering
\begin{tikzpicture}[
    node distance=4mm,
    box/.style={draw, rounded corners=2pt, minimum height=8mm, minimum width=14mm, align=center, font=\scriptsize, inner sep=2pt},
    neural/.style={box, fill=blue!10},
    symbolic/.style={box, fill=orange!15},
    output/.style={box, fill=green!15},
    input/.style={box, fill=gray!10},
    arr/.style={-{Latex[length=1.5mm]}, thick}
]
\node[input] (t)   {Transcript\\+ Disposition};
\node[neural, right=of t]   (p) {SOP\\Prompt};
\node[neural, right=of p]   (m) {LLaMA 8B\\(LoRA)};
\node[symbolic, right=of m] (r) {Rule-Based\\Post-Proc.};
\node[output, right=of r]   (o) {Corrected\\JSON};
\draw[arr] (t) -- (p);
\draw[arr] (p) -- (m);
\draw[arr] (m) -- node[above, font=\tiny] {raw} (r);
\draw[arr] (r) -- (o);
\draw[dashed, ->] (t.south) .. controls +(0,-0.6) and +(0,-0.6) .. (r.south);
\node[font=\scriptsize\bfseries, above=1mm of m, color=blue!60!black] {Stage 1: Neural};
\node[font=\scriptsize\bfseries, above=1mm of r, color=orange!70!black] {Stage 2: Symbolic};
\end{tikzpicture}
\caption{Hybrid inference architecture. Stage 1 generates structured JSON; Stage 2 enforces deterministic SOP rules over the original transcript.}
\label{fig:arch}
\end{figure}
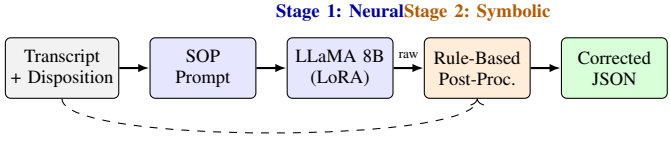

Fig.~\ref{fig:arch} shows the two-stage hybrid inference pipeline. The fine-tuned model generates a structured JSON evaluation (Stage 1), which is then validated and corrected by deterministic SOP rules applied over the original transcript (Stage 2).

\subsection{Multi-Stage Data Curation}
\textbf{Stage 1: Noise removal} (209 $\to$ 199). Removed 10 examples with encoding artifacts, fragments $<$30 characters, or system errors.

\textbf{Stage 2: Hard-negative augmentation} (199 $\to$ 219). Agent Disclosure requires the exact phrase $\phi^{*}$. Near-miss variants (e.g., omitting ``AI'' or substituting ``automated'') must be classified as failures. We synthesized 20 adversarial examples from 10 near-miss templates across three violation categories:
\begin{itemize}
    \item \emph{Partial omission}: ``I am a Collections agent'' (missing ``AI'')
    \item \emph{Synonym substitution}: ``I am an automated Collections agent''
    \item \emph{Complete omission}: ``I am here to help you''
\end{itemize}
Each hard negative preserves the full transcript context but modifies only the disclosure phrase, creating minimally contrastive pairs with opposite labels.

\textbf{Stage 3: Exhaustive SOP audit} (219 examples). Automated validation of all fields: regex-based amount extraction, authorization pattern matching, disposition-outcome consistency, and cross-field constraint enforcement per Eq.~(1).

\begin{table}[t]
\centering
\caption{Training Configuration.}
\label{tab:train}
\begin{tabular}{ll}
\toprule
\textbf{Parameter} & \textbf{Value} \\
\midrule
Base model & LLaMA 3.1 8B Instruct \\
Precision & bfloat16 (no quantization) \\
LoRA rank / alpha / dropout & 64 / 128 / 0.05 \\
Target modules & All 7 projections \\
Trainable params & 167.8M / 8.2B (2.05\%) \\
Hardware & 4$\times$ NVIDIA H100 80GB \\
Epochs & 8 (early stop patience=4) \\
Learning rate & $1.5 \times 10^{-4}$, cosine \\
Effective batch size & 32 (4 GPU $\times$ 4 $\times$ 2) \\
Optimizer / weight decay & AdamW / 0.05 \\
\bottomrule
\end{tabular}
\end{table}

\subsection{LoRA Fine-Tuning}
For pre-trained weight $W_0 \in \mathbb{R}^{d \times k}$, LoRA constrains the update:
\begin{equation}
W = W_0 + \Delta W = W_0 + BA
\end{equation}
where $B \in \mathbb{R}^{d \times r}$, $A \in \mathbb{R}^{r \times k}$, $r \ll \min(d, k)$. The forward pass:
\begin{equation}
h = W_0 x + \frac{\alpha}{r} BA x
\end{equation}

We apply LoRA to all seven projection matrices $\{W_q, W_k, W_v, W_o, W_{\text{gate}}, W_{\text{up}}, W_{\text{down}}\}$ in each of the 32 transformer layers with $r = 64$, $\alpha = 128$, yielding 167.8M trainable parameters (2.05\% of 8.2B).

\subsection{Training Configuration}
Table~\ref{tab:train} summarizes the training setup. We use label masking to compute loss only on assistant (JSON output) tokens:
\begin{equation}
\mathcal{L} = -\frac{1}{|\mathcal{A}|} \sum_{t \in \mathcal{A}} \log p_\theta(x_t \mid x_{<t})
\end{equation}
where $\mathcal{A}$ is the set of assistant token positions and $\theta$ denotes LoRA parameters. System/user tokens are masked with label $-100$.

\subsection{Deterministic Post-Processing}
The post-processing layer $g(\tilde{\mathbf{y}}, T) \to \hat{\mathbf{y}}$ enforces five rules:

\noindent\textbf{R1.} Agent Disclosure: exact substring match for $\phi^{*}$, overriding the model unconditionally.

\noindent\textbf{R2.} Payment Disclosure: conjunction of authorization script detection (5 patterns) and post-authorization verbal confirmation.

\noindent\textbf{R3.} Disposition: bankruptcy dispositions $\to$ Fail; transfer-disposition consistency enforcement.

\noindent\textbf{R4.} Compliance Result: recomputed via Eq.~(1).

\noindent\textbf{R5.} Feedback regeneration when corrections are applied.

This decomposition assigns contextual reasoning (professionalism assessment, payment intent inference) to the neural model and deterministic rules (exact matching, cross-field arithmetic) to the symbolic layer.

\section{Experimental Setup}
\subsection{Dataset}
219 labeled transcripts from a production conversational AI system in financial services. Split: 188 train / 21 eval (90/10, seed=42). Token lengths: min 1{,}067, max 2{,}472, mean 1{,}350. Compliance distribution: 83 Pass (39.7\%) / 126 Fail (60.3\%). 20 hard-negative augmented examples (9.1\% of dataset).

\subsection{Blind Test Set}
53 production transcripts completely unseen during training, each independently reviewed by a human compliance expert who marked predictions as ``Correct'' or ``Incorrect'' with error annotations.

\subsection{Deployment}
Inference on a single NVIDIA A100 80GB GPU (\$134/day). Greedy decoding (temperature=0, repetition penalty=1.1, max 1024 tokens).

\section{Results}
\subsection{Training Dynamics}
\begin{figure}[t]
\centering
\begin{tikzpicture}
\begin{axis}[
    width=0.95\columnwidth, height=5cm,
    xlabel={Training Step}, ylabel={Loss},
    xmin=10, xmax=100, ymin=0, ymax=0.35,
    legend pos=north east,
    legend style={font=\scriptsize},
    grid=both, grid style={gray!20},
    tick label style={font=\scriptsize},
    label style={font=\scriptsize}
]
\addplot[blue, mark=*, thick] coordinates {
    (10,0.322) (20,0.208) (30,0.155) (40,0.075) (50,0.070)
    (60,0.073) (70,0.052) (80,0.034) (90,0.029) (100,0.017)
};
\addlegendentry{Train}
\addplot[red, mark=square*, thick] coordinates {
    (10,0.219) (20,0.175) (30,0.142) (40,0.135) (50,0.138)
    (60,0.1305) (70,0.138) (80,0.146) (90,0.137) (100,0.152)
};
\addlegendentry{Val}
\addplot[green!60!black, mark=triangle*, mark size=4pt, only marks] coordinates {(60,0.1305)};
\addlegendentry{Best}
\node[font=\tiny, green!50!black] at (axis cs:60,0.165) {0.1305};
\end{axis}
\end{tikzpicture}
\caption{Training and validation loss. Best checkpoint at step 60 (val loss 0.1305). Train-val divergence after step 60 indicates overfitting on 188 examples.}
\label{fig:loss}
\end{figure}
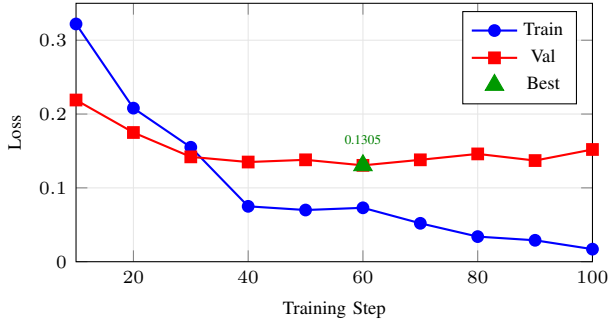

Fig.~\ref{fig:loss} shows the loss curves. Training loss drops from 0.322 to 0.017 (94.7\% reduction) while validation loss bottoms at 0.1305 (step 60) before rising. The $\sim$9$\times$ train-val gap at step 100 confirms that LoRA's parameter efficiency and early stopping are both critical for this low-resource regime.

\subsection{Blind Test Results}
Table~\ref{tab:results} presents the main results on 53 blind production transcripts.

\begin{table}[t]
\centering
\caption{Per-Field Accuracy on 53 Blind Test Examples (Human-Validated).}
\label{tab:results}
\begin{tabular}{llr}
\toprule
\textbf{Type} & \textbf{Field} & \textbf{Accuracy} \\
\midrule
\multirow{4}{*}{I -- Critical}
 & Agent Disclosure   & \textbf{100.0\%} \\
 & Payment Disclosure & 96.2\% \\
 & Disposition        & 90.6\% \\
 & Compliance Result  & 90.6\% \\
\midrule
\multirow{5}{*}{II -- Behavioral}
 & Threats            & 100.0\% \\
 & Transfer Accuracy  & 100.0\% \\
 & Incorrect Payment  & 100.0\% \\
 & Prohibited Payment & 100.0\% \\
 & Unprofessionalism  & 98.1\% \\
\midrule
III -- Info & Documentation & 100.0\% \\
\midrule
\textbf{JSON Structural Validity} & & \textbf{100.0\%} \\
\textbf{Overall (human-validated)} & & \textbf{83.0\%} \\
Inference Latency (1$\times$A100) & & $\sim$2 sec \\
\bottomrule
\end{tabular}
\end{table}

Key findings: (1) Agent Disclosure achieves 100\%, validating hard-negative augmentation. (2) JSON validity is 100\%, versus typical 85--95\% for zero-shot frontier models. (3) Disposition (90.6\%) is the primary error source, with 5 of 9 errors being false positives.

\subsection{Error Analysis}
The model predicted Disposition=Pass for all 53 cases, never predicting Fail---revealing training distribution bias (67\% correct dispositions). Targeted augmentation analogous to the successful Agent Disclosure hard negatives would likely address this.

\begin{table}[t]
\centering
\caption{Error Breakdown (9 Errors across 53 Evaluations).}
\label{tab:errors}
\begin{tabular}{lrr}
\toprule
\textbf{Error Type} & \textbf{Count} & \textbf{\%} \\
\midrule
Disposition false positive  & 5 & 55.6 \\
Payment Disclosure miss     & 2 & 22.2 \\
Payment context miss        & 1 & 11.1 \\
Unprofessionalism miss      & 1 & 11.1 \\
\bottomrule
\end{tabular}
\end{table}

\subsection{Cost and Latency Analysis}
Table~\ref{tab:cost} demonstrates 46--76\% cost reduction versus per-transcript API alternatives and $>$97\% versus in-context learning. Monthly savings versus GPT-4o: \$3{,}480.

\begin{table}[t]
\centering
\caption{Cost and Latency at 10K Evaluations/Day. GPU: \$134/day. API pricing from~\cite{artificialanalysis}.}
\label{tab:cost}
\begin{tabular}{lrrrc}
\toprule
\textbf{Approach} & \textbf{Latency} & \textbf{Per-Eval} & \textbf{Monthly} & \textbf{Data On-Prem} \\
\midrule
\textbf{Ours (A100)} & \boldmath$\sim$\textbf{2s} & \textbf{\$0.013} & \textbf{\$4{,}020} & \checkmark \\
GPT-4o API           & $\sim$4s  & \$0.025 & \$7{,}500  & $\times$ \\
Claude Sonnet API    & $\sim$6s  & \$0.033 & \$9{,}900  & $\times$ \\
Claude Opus API      & $\sim$10s & \$0.055 & \$16{,}500 & $\times$ \\
Few-Shot Prompting   & & & & \\
\quad (250K token context) & 30--90s & \$0.65+ & \$195K+ & $\times$ \\
\bottomrule
\end{tabular}
\end{table}

\subsection{Impact of Hard-Negative Augmentation}
The 100\% accuracy on Agent Disclosure---versus 90.6\% on Disposition (which lacked targeted augmentation)---provides direct evidence that hard-negative augmentation at critical decision boundaries is highly effective. Just 20 synthetic examples (9.1\% of the dataset) eliminated all disclosure classification errors.

\section{Discussion}
\subsection{Small Models vs. Frontier APIs}
Three mechanisms explain why our 8B-parameter model matches frontier models on this task:

\textbf{Weight-encoded knowledge.} Fine-tuning encodes domain rules in model weights, accessible at $O(1)$ cost per inference. In-context learning encodes rules in the context window, subject to attention degradation~\cite{liu2024lost}.

\textbf{Schema consistency.} Fine-tuning on structured JSON with label masking teaches the exact output format, achieving 100\% JSON validity versus 85--95\% for zero-shot approaches.

\textbf{Decision boundary calibration.} Hard-negative augmentation directly calibrates boundaries for critical rules---impossible with zero-shot prompting alone.

\subsection{The Hybrid Decomposition Principle}
Our architecture embodies a general principle for regulated domains: \emph{decompose tasks into contextual reasoning (neural) and constraint enforcement (symbolic)}. The neural model need not be perfect on every field---deterministic fields are corrected at zero cost. This concentrates model capacity on genuinely difficult subproblems requiring transcript understanding.

\subsection{Limitations}
\textbf{Statistical power.} 53-example blind test yields $\pm$10\% confidence intervals at 95\% level.

\textbf{Disposition bias.} The model never predicted Disposition=Fail, revealing training distribution bias addressable via targeted augmentation.

\textbf{Class imbalance.} Extreme imbalance (Threats: 98\% Pass) limits minority-class evaluation.

\textbf{Post-processing brittleness.} Regex-based rules may fail on novel conversational patterns.

\section{Conclusion}
We demonstrated that a domain-adapted small language model (LLaMA 3.1 8B, 2.05\% trainable via LoRA) fine-tuned on only 219 examples, combined with deterministic post-processing, achieves 83\% human-validated accuracy on multi-label compliance evaluation---with 100\% JSON validity, $\sim$2-second latency, and \$0.013 per evaluation. Compared to frontier API alternatives, this represents 2--5$\times$ latency reduction, 46--76\% cost savings, and full data privacy through on-premise deployment.

Our hard-negative augmentation strategy achieves 100\% accuracy on the most critical field using just 20 synthetic examples, demonstrating that targeted data curation can substitute for model scale. The hybrid neural-deterministic architecture provides a general blueprint for deploying language models in regulated industries where both contextual understanding and strict rule compliance are required.


\begin{thebibliography}{99}
\bibitem{brown2020} T. Brown et al., ``Language models are few-shot learners,'' in \emph{Advances in NeurIPS}, vol. 33, pp. 1877--1901, 2020.
\bibitem{hu2022lora} E. J. Hu et al., ``LoRA: Low-rank adaptation of large language models,'' in \emph{Proc. ICLR}, 2022.
\bibitem{dubey2024llama3} A. Dubey et al., ``The LLaMA 3 herd of models,'' \emph{arXiv:2407.21783}, 2024.
\bibitem{liu2024lost} N. F. Liu et al., ``Lost in the middle: How language models use long contexts,'' \emph{TACL}, vol. 12, pp. 157--173, 2024.
\bibitem{dettmers2023qlora} T. Dettmers et al., ``QLoRA: Efficient finetuning of quantized language models,'' in \emph{Advances in NeurIPS}, vol. 36, 2023.
\bibitem{cui2023chatlaw} J. Cui et al., ``ChatLaw: Open-source legal large language model,'' \emph{arXiv:2306.16092}, 2023.
\bibitem{singhal2023clinical} K. Singhal et al., ``Large language models encode clinical knowledge,'' \emph{Nature}, vol. 620, pp. 172--180, 2023.
\bibitem{wu2023bloomberggpt} S. Wu et al., ``BloombergGPT: A large language model for finance,'' \emph{arXiv:2303.17564}, 2023.
\bibitem{wang2023selfinstruct} Y. Wang et al., ``Self-instruct: Aligning language models with self-generated instructions,'' in \emph{Proc. ACL}, 2023.
\bibitem{willard2023guided} B. T. Willard and R. Louf, ``Efficient guided generation for large language models,'' \emph{arXiv:2307.09702}, 2023.
\bibitem{gao2021simcse} T. Gao, X. Yao, and D. Chen, ``SimCSE: Simple contrastive learning of sentence embeddings,'' in \emph{Proc. EMNLP}, pp. 6894--6910, 2021.
\bibitem{feng2021augsurvey} S. Y. Feng et al., ``A survey of data augmentation approaches for NLP,'' in \emph{Findings of ACL}, pp. 968--988, 2021.
\bibitem{artificialanalysis} Artificial Analysis, ``Independent LLM benchmarks: Speed, quality, and price,'' \url{https://artificialanalysis.ai}, 2024.
\bibitem{touvron2023llama} H. Touvron et al., ``LLaMA: Open and efficient foundation language models,'' \emph{arXiv:2302.13971}, 2023.
\bibitem{vaswani2017attention} A. Vaswani et al., ``Attention is all you need,'' in \emph{Advances in NeurIPS}, vol. 30, 2017.
\end{thebibliography}
\end{document}